\documentclass[sigconf]{acmart}

\AtBeginDocument{%
  \providecommand\BibTeX{{%
    \normalfont B\kern-0.5em{\scshape i\kern-0.25em b}\kern-0.8em\TeX}}}

\renewcommand\footnotetextcopyrightpermission[1]{} 

\acmSubmissionID{1865}

\usepackage{cleveref}
\usepackage{multirow}
\usepackage{listings}
\usepackage[normalem]{ulem}
\usepackage{subcaption} 
\usepackage{colortbl}
\usepackage{booktabs}    
\usepackage{multirow}    

\newcommand{\rra}[1]{RRA}
\newcommand{\mdsfull}[1]{Mask Guided Diffusion Process}
\newcommand{\rrafull}[1]{Residual Reference Attention}
\newcommand{\fdtefull}[1]{Frequency-aware Decoupled Textual Embedding}

\begin{document}

\title{FocusDiT: Masking Queries in Diffusion Transformers for \\ Fine-grained Image Generation}


\author{Xueji Fang}
\affiliation{
  \institution{Zhejiang University \\
  Westlake University}
  \city{Hangzhou}
  \country{China}
  }
\email{fangxueji@zju.edu.cn}

\author{Liyuan Ma\footnotemark[1]}
\thanks{*Corresponding author.}
\affiliation{%
  \institution{Westlake University}
  \city{Hangzhou}
  \country{China}
  }
\email{maliyuan@westlake.edu.cn}

\author{Jianhao Zeng}
\affiliation{%
  \institution{Westlake University}
  \city{Hangzhou}
  \country{China}
  }
\email{zengjianhao@westlake.edu.cn}

\author{Jinjin Cao}
\affiliation{%
  \institution{Zhejiang University \\
  Westlake University}
  \city{Hangzhou}
  \country{China}
  }
\email{caojinjin@westlake.edu.cn}

\author{Mingyuan Zhou}
\affiliation{%
  \institution{Westlake University}
  \city{Hangzhou}
  \country{China}
  }
\email{zhoumingyuan@westlake.edu.cn}

\author{Guo-Jun Qi\footnotemark[1]}
\affiliation{%
  \institution{Westlake University}
  \city{Hangzhou}
  \country{China}
  }
\email{guojunq@gmail.com}

\renewcommand{\shortauthors}{ Xueji Fang, Liyuan Ma, and Guo-Jun Qi}


\begin{abstract}
Diffusion transformer (DiT) has been widely adopted in the generative diffusion field, advancing the denoising of query tokens through attention and Feed-Forward (\text{FFN}) layers. 
FFN actually acts as the key-value vocabulary for decoding visual contents where the value embeds the visual semantical knowledge.
We present that focusing on critical query tokens corresponding to more complex details and encouraging the model to improve these tokens is essential for fine-grained visual generation.
To this end, we propose FocusDiT, which applies a Masking scheme to focus on critical query tokens that are exclusively fed into FFN. 
The masked queries can retrieve visual tokens from the FFN vocabularies, and use them  to decode their visual details.
Extensive text-to-image experiments validate the effectiveness of token masking in enhancing generative performance.

 \end{abstract}


\begin{teaserfigure}
  \includegraphics[width=\textwidth]{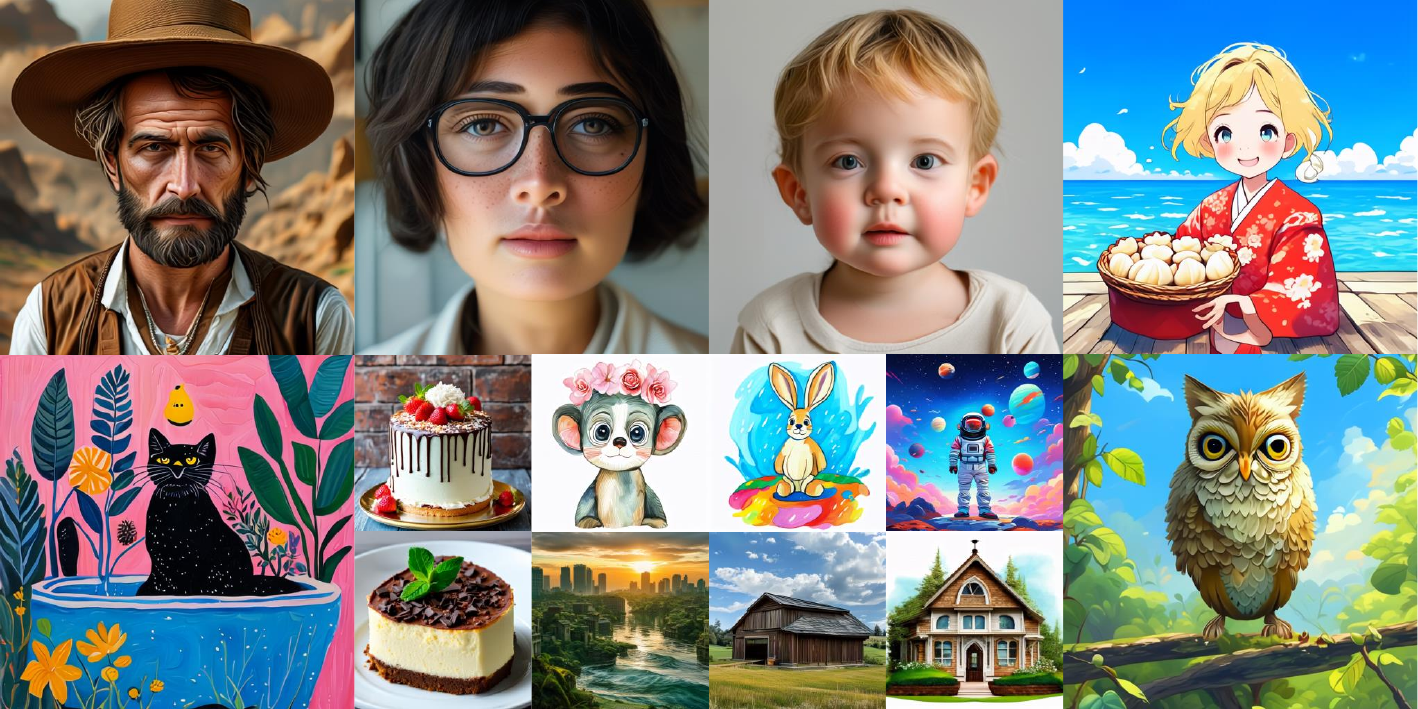}
  \captionof{figure}{Fine-grained text-to-image samples from our FocusDiT, showcasing its capabilities in attention to fine details and high image quality in various styles.}
  \label{fig:teaser}
\end{teaserfigure}

\maketitle

\section{Introduction}
\label{sec:intro}


\newcommand{\zjh}[1]{\textcolor{red}{#1}}
\newcommand{\fxj}[1]{\textcolor{green}{#1}}


Diffusion transformers~\cite{DiT} have become the state-of-the-art approach in visual generation, showing impressive performance in various tasks.
An increasing amount of research has been devoted into the design of diffusion transformer architectures. 
For instance, DiG~\cite{DiG} introduces linear attention to address computational load issues, while Stable Diffusion 3~\cite{SD3} enhances text-visual alignment by incorporating a multi-branch structure for text and visual interaction within the attention mechanism. 
Several studies also target the FFN, such as EC-DiT~\cite{ECDiT}, which introduces Mixture-of-Experts (MoE) techniques into the DiT, scaling up the model by increasing the number of FFNs to improve performance. 
Prior analyses of transformer FFNs show that FFN layers can act as key-value memories and promote concepts through a vocabulary-like intermediate space~\cite{key-value, Build-Predictions}.
However, there is still a lack of comprehensive study and optimization of the FFN in DiT that plays a key role in decoding visual content.

\begin{figure}[t]
  \centering
  \includegraphics[width=\linewidth]{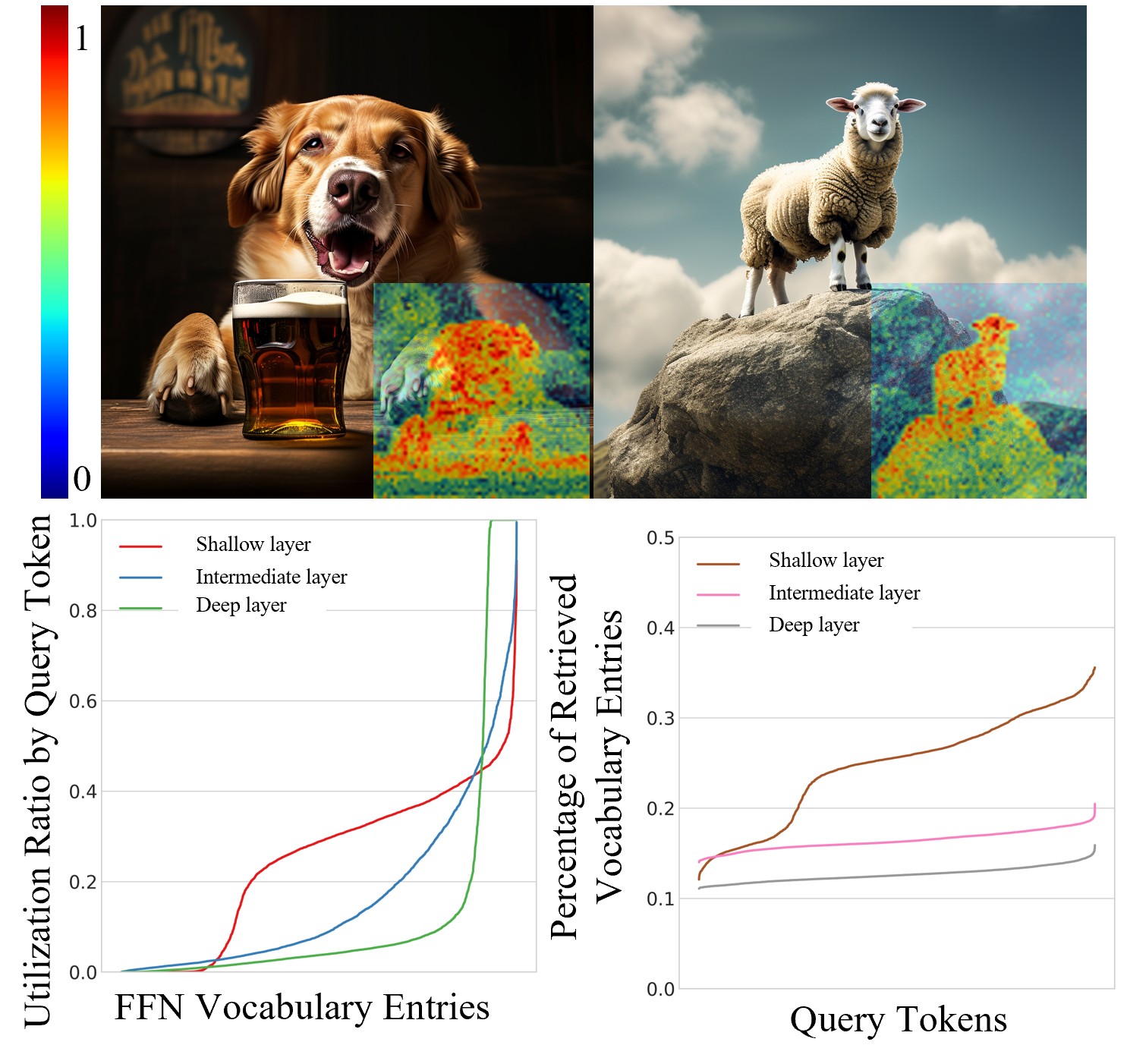}
   \caption{
   \textbf{Top}: Heatmaps indicating the number of entries from certain vocabulary subset retrieved by query tokens from different content, highlighting that certain vocabulary tokens correspond to specific semantic (e.g., animals).
   \textbf{Bottom}: Analysis of vocabulary utilization across layers and query tokens, based on 100 samples from DiT-based model~\cite{Pixart-alpha}.
    }
   \label{fig:insight}
\end{figure}

In large transformer models, the FFN has been shown to function as a key-value vocabulary, and the DiT leverages the FFN to store visual tokens required for generating diverse visual semantics. 
These visual tokens, representing the visual semantical knowledge learned from the training data  as presented in the left of ~\autoref{fig:insight}, are embedded into the FFN's weights and are subsequently accessed by the \emph{query tokens} from preceding attention layers.
We argue that critical query tokens characterizing fine-grained visual details should be selectively masked and fed into the FFN to retrieve more visual tokens for visual content decoding.
However, the critical tokens are often overwhelmed by those corresponding to backgrounds or low-frequency regions of few visual details, wasting the full utilization of FFN's visual vocabulary to refine less important query tokens.
This interference in utilization across FFN vocabulary hinders further performance improvement. 
Additionally, the FFN vocabulary is not fully leveraged as indicated in the right of ~\autoref{fig:insight}, with certain elements in shallow and deep layers seldom being utilized by the query tokens.
This suggests that the decoding process in different FFN layers requires vocabularies of varying sizes.

To address these challenges, we propose \textbf{FocusDiT} which applies \emph{Query Token Masking} to boost the FFN vocabulary utilization for critical tokens.
The query mask distinguishes between tokens with complex structures and those with simpler ones, masking out the latter to avoid interference in decoding the visual content of critical tokens.
Besides, we adjust the vocabulary organization of FFNs through a local-expert and shared-expert design, which preserves the per-layer FFN computation while reducing redundant layer-specific parameters.
In particular, each FFN keeps a local vocabulary for layer-specific decoding and reuses a shared vocabulary across layers, allowing frequently needed visual concepts to be represented without manually designing a layer-wise vocabulary schedule.
Further investigation  shows that with accurate query mask selection, FocusDiT enhances fine-grained visual generation. 
Additionally, the query mask improves inference efficiency by skipping FFN layers where most mask values are close to zero.
Both quantitative and qualitative results from text-to-image experiments show advantages over competing methods, confirming the effectiveness of our approach.
Our contributions are summarized as follows:
\begin{itemize}
    \item We propose a Query Token Masking strategy that identifies and prioritizes critical query tokens containing complex structures, enhancing vocabulary utilization for these tokens to capture fine-grained details.
    Besides, the query mask can be used to decrease inference cost, thereby enriching its practical significance.
    \item We propose a Vocabulary Redistribution (VR) scheme with local and shared FFN experts, reducing redundant FFN parameters while preserving per-layer decoding capacity and improving the robustness of the vocabulary design.
    \item Extensive experiments and analyses on visual generation tasks demonstrate the superiority of FocusDiT in both visual quality and quantitative evaluation.
    
\end{itemize}

\section{Related Work}
\label{sec:relatedwork}

\begin{figure*}[htpb]
    \centering
    \includegraphics[width=\linewidth]{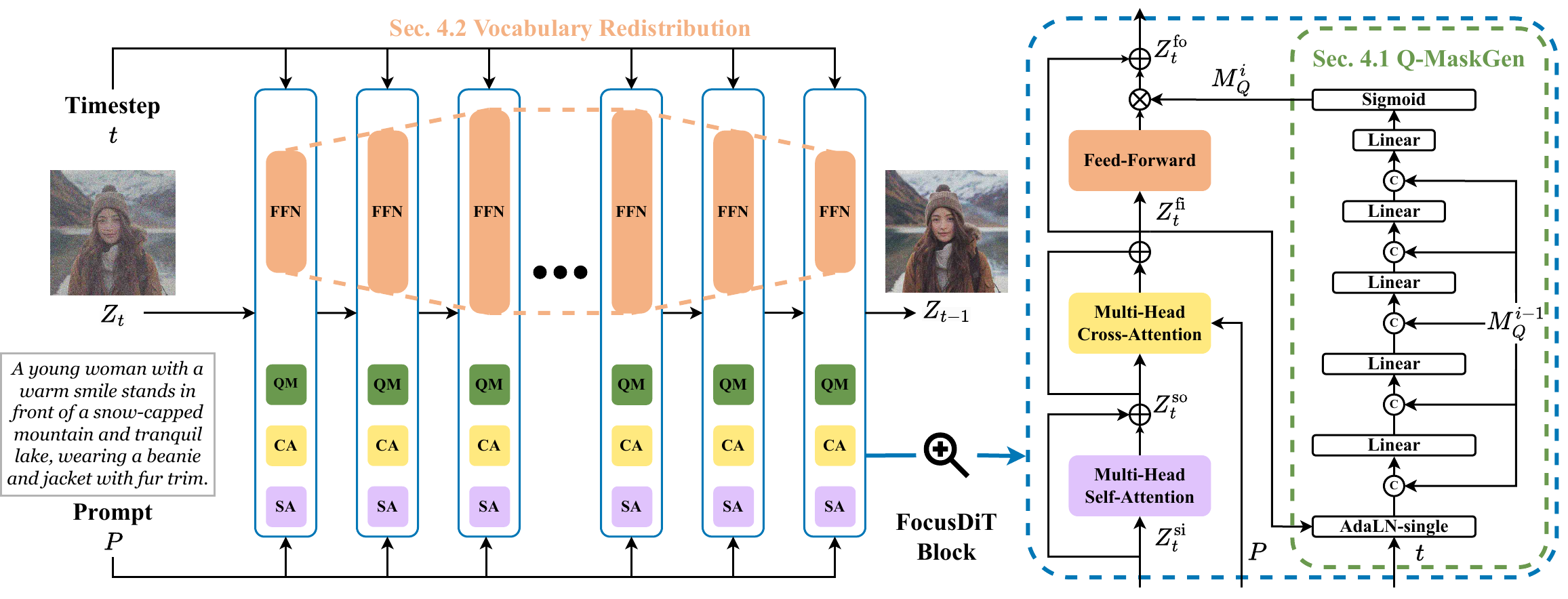}
    \caption{Framework of our proposed FocusDiT. \textbf{Left:} Overview of the main architecture, including multiple FocusDiT blocks. \textbf{Right:} Architecture of our query mask prediction network (Q-MaskGen).}
    \label{fig:framework}
\end{figure*}
\subsection{Generative Diffusion Models}
Diffusion models~\cite{DDPMs,IDDPM,DDIM,controlnet,EDM,FlowMatching, diffusionsurvey} have shown impressive performance in generative tasks, trained with denoising objective $\mathcal{L}_{DM} = \mathbb{E}_{x_0, \theta, t}\left [ \left \Vert \epsilon	-\epsilon_\theta(\sqrt{\bar{\alpha}}_t x_0+\sqrt{1-\bar{\alpha}_t}\epsilon, t) \right \Vert_2^2 \right ]$ with $\epsilon\sim\mathcal{N}\left ( 0, \mathbf{I} \right )$, timestep $t$, dataset sample $x_0$, timestep-related hyperparameter $\bar{\alpha}_t$, model parameters $\theta$.
DiT replaces the U-Net~\cite{UNet} backbone with transformer for a range of generative diffusion modeling tasks, yielding notable improvements in both performance and scalability. 
For instance, PixArt series~\cite{Pixart-alpha, Pixart-sigma, Pixart-delta} incorporates a cross-attention module for the text-to-image task, while Sora~\cite{Sora} further extends DiTs for text-to-video generation.
Most improvements to DiTs have focused on optimizing the attention module~\cite{DiG, SD3, flux1_dev, flux1_schnell}, with comparatively few studies exploring the enhancement of the FFN.
Earlier approaches~\cite{ECDiT, DiT-MoE} explored employing MoE to integrate multiple FFN expert modules, expanding the model without altering the internal structure of the FFN.
However, these methods have not deeply analyzed how the internal mechanisms of the FFN affect the generative process, which is the focus of our work.

\subsection{FFN in Large Transformer Model}
Research on large transformer model~\cite{few-shot-comet, LPAQA, knowledge-neurons, Kformer, rome, BERTnesia, key-value, Build-Predictions} has shown that FFNs serve as the knowledge base for storing information from training dataset, retained in a key-value memory format during model training, enabling specific knowledge to be activated by input and mapped to the final generated output. 
Inspired by this, we examine the role of the FFN in DiT, which shares structural similarities to LLMs within the visual content generation process, finding that the FFN also acts as a vocabulary, storing various types of semantic concepts. 
Different from FFN-level scaling methods that add more experts or increase model size, FocusDiT studies how query tokens activate the existing FFN vocabulary and strengthens this activation through a learned mask.
The proposed Vocabulary Redistribution further reduces redundant FFN parameters by sharing part of the vocabulary across layers, while keeping the FFN computation of each block unchanged.

\subsection{Token Selection}
Token selection mechanisms are common adopted in generative visual and language modeling, enabling models to selectively distinguish and leverage tokens based on their importance~\cite{Attention,TNT,Mixformerv2,BST}.
For instance, DynamicViT~\cite{Dynamicvit} selectively prunes less informative tokens to improve the model's efficiency. DiffRate~\cite{Diffrate} employs a token importance metric to identify and select the top-K tokens through token pruning and merging. 
DyDiT~\cite{DyDiT} dynamically selects well-denoised tokens and bypasses subsequent blocks, thereby reducing the model's computational load. 
These works mainly aim to improve the model's operational efficiency.
In contrast, our query masking strategy directs the model to prioritize the vocabulary allocation to critical tokens, thereby improving visual generation quality.
The learned soft Q-Mask is also different from heuristic layer skipping or token reduction: it is applied to FFN updates during training to improve token decoding, while its binarized or thresholded form can optionally be used for inference acceleration.
Thus, FocusDiT is complementary to token pruning, token merging, and layer-skipping methods rather than a replacement for them.




\section{Preliminary}
FFN in DiT actually serves as key-value vocabulary which stores the learned visual semantics~\cite{key-value, Build-Predictions}.
Specifically, FFN consists of two weight matrices $W_{K}\in \mathbb{R}^{d\times d_m}$, $W_{V}\in \mathbb{R}^{d_m\times d}$ and corresponding biases $b_K \in \mathbb{R}^{d_m}$, $b_V \in \mathbb{R}^{d}$.
$d_m$ actually represents the vocabulary size.
The input query token $Z^{\text{fi}}_t\in \mathbb{R}^d$ interacts with the first weight matrix corresponding to \emph{key} and then aggregates vocabulary entries saved in second \emph{value} weight matrix $W_V$ to produce final output $Z^{\text{fo}}_t$.
$\text{FFN}(\cdot)$ calculation is denoted as:
\begin{equation}
    \begin{aligned}
        Z^{\text{fo}}_t  &= Z^{\text{fi}}_t + \text{FFN}(Z^{\text{fi}}_t) \\
        &= Z^{\text{fi}}_t + [f(Z^{\text{fi}}_t\cdot {W_K} + b_K)\cdot W_{V} + b_V],
    \end{aligned}
    \label{eq:ffn}
\end{equation}
where $f$ is a non-linearity activation function such as GeLU. 

\noindent\textbf{Vocabulary utilization in FFN is insufficient and not distinguished for DiT.}
We quantify FFN behavior by treating the output of the first FFN projection and activation as a vocabulary activation map.
For an FFN input with shape $B \times N \times C_1$, the first linear layer and GeLU activation produce $X \in \mathbb{R}^{B \times N \times C_2}$, where $X_{b,n,c}$ denotes the activation of the $c$-th vocabulary entry for the $n$-th query token in batch $b$.
A vocabulary entry is considered activated when its value is positive.
We define the utilization ratio $R_{\text{utilize}}$ as the proportion of vocabulary entries activated by each query token, and the retrieval ratio $R_{\text{retrieve}}$ as the proportion of query tokens that activate each vocabulary entry:
\begin{lstlisting}[language=Python, numbers=none, basicstyle=\ttfamily\small, columns=fullflexible, breaklines=true]
# Start FFN
X = FFN.linear1(X)                         # B*N*C1 -> B*N*C2
X = FFN.GeLU(X)
R_utilize = (X > 0).sum(dim=-1) / X.shape[-1]  # B*N
R_retrieve = (X > 0).sum(dim=-2) / X.shape[-2] # B*C2
X = FFN.linear2(X)
# End FFN
\end{lstlisting}
In~\autoref{fig:insight}, the horizontal axes of the vocabulary entries and query tokens are sorted by their corresponding ratios for clearer visualization.
As shown in the bottom left of~\autoref{fig:insight}, vocabulary utilization is insufficient, with many entries rarely activated during the generation process.
Although DiT can access relevant visual tokens from the FFN for denoising, the allocation to query tokens shows minimal variation as shown in the bottom right of~\autoref{fig:insight}, failing to distinguish between critical and non-critical tokens, which limits the ability of critical tokens to access more vocabulary entries.
These limitations hinder the effective utilization of vocabulary and degrade generation quality.


\section{Methodology}
\label{sec:methodology}
\begin{figure*}[htpb]
    \centering
    \includegraphics[width=\linewidth]{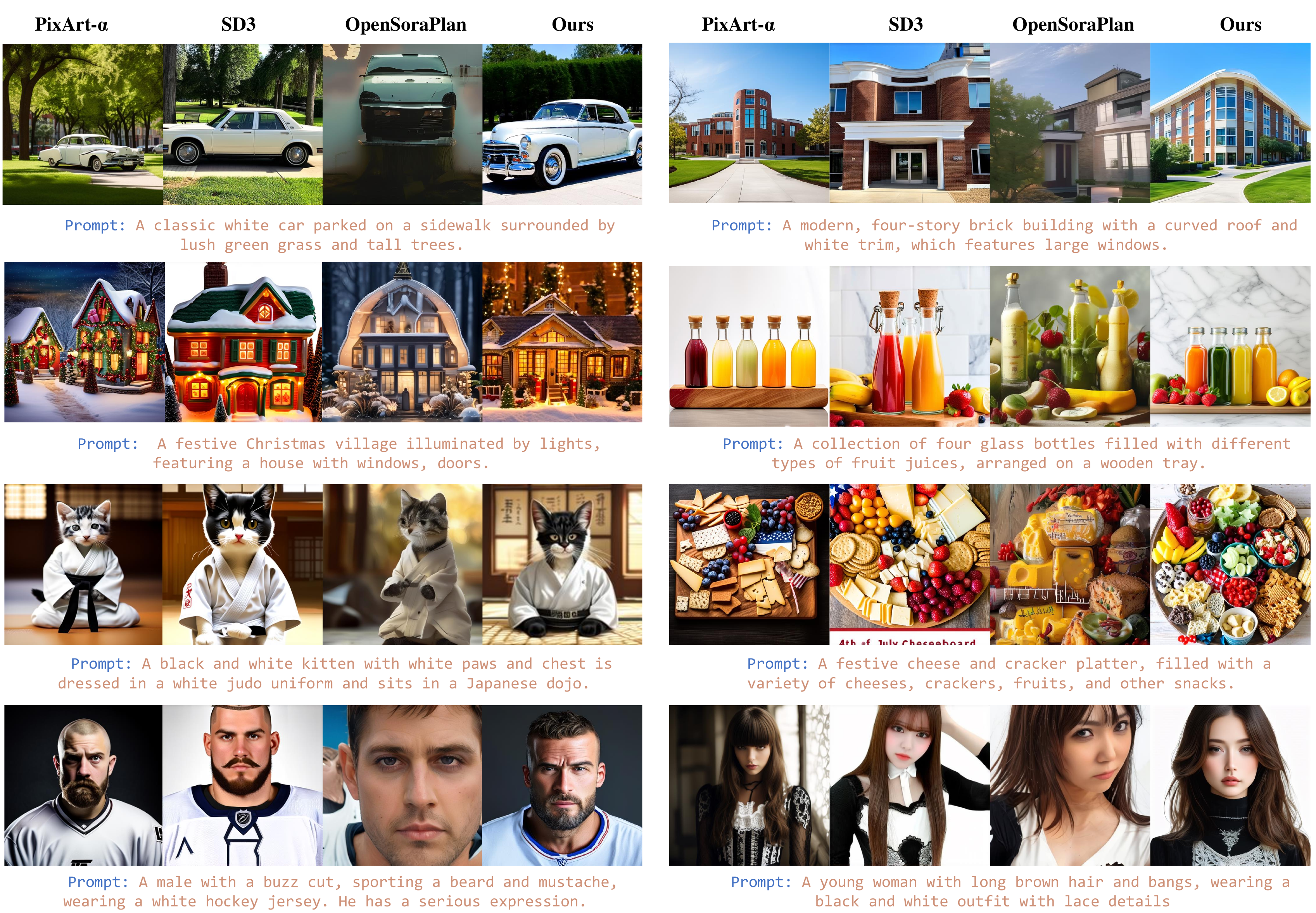}
    \caption{Qualitative comparison with PixArt-$\alpha$, SD3 and OpenSoraPlan.}
    \label{fig:quali}
\end{figure*}
Based on the analyses above and with the aim of enhancing critical query token decoding, we formulate the hypothesis: \emph{\textbf{Can giving identified query tokens a greater share of the FFN vocabulary improve generation outcomes?}}
To explore this, we design Query Mask (Section~\ref{sec:query_mask}) to prioritize the critical query token generation and Vocabulary Redistribution (Section~\ref{sec:vr}) to adjust the vocabulary size for more effective query token decoding.


\subsection{Query Mask for Token Focus}
\label{sec:query_mask}
\begin{figure*}[htpb]
    \centering
    \includegraphics[width=0.95\linewidth]{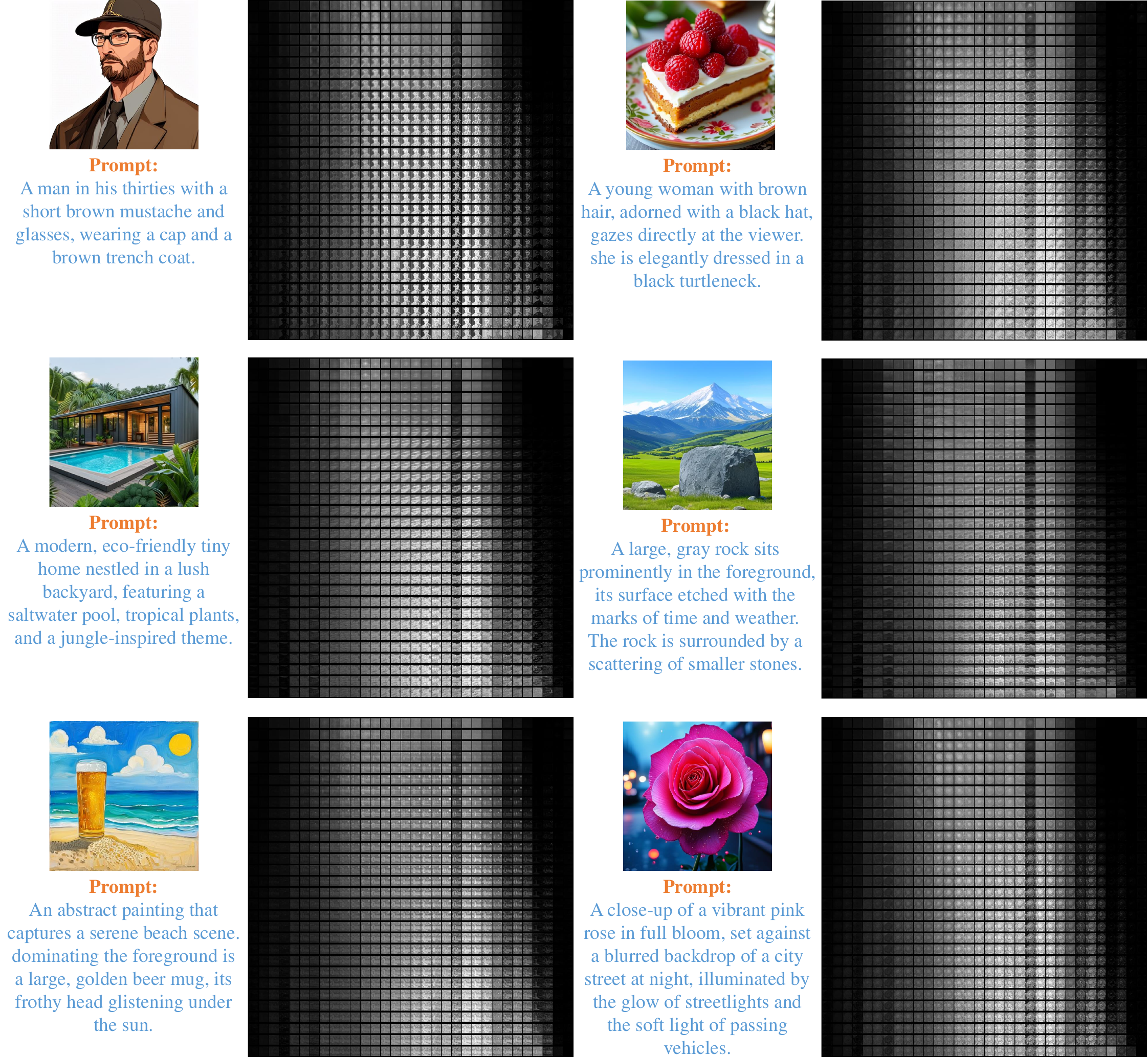}
    \caption{Visualization of query masks across different FocusDiT blocks and denoising timesteps. From left to right, the grid represents the progression from the first Transformer block to the last, while from top to bottom, it corresponds to increasing denoising timesteps. Lighter regions highlight critical query tokens that play a more significant role in decoding, whereas darker regions indicate less influential tokens.}
    \label{fig:mq_vis}
\end{figure*}

The query mask is designed to guide critical tokens to query richer visual semantics from the FFN vocabulary to improve their decoded representation. 
This ensures that the model can focus on decoding query tokens for various visual details.
To achieve this, the query mask dynamically predicts which tokens are responsible for complex details, marking them with a value close to 1.

The prediction of the query mask considers the diffusion timestep, since the query tokens that require focus vary across noise levels.
At high noise levels, the information across query tokens is relatively uniform, so no special distinction is needed. 
However, as the noise level decreases, critical query tokens that represent complex details become harder to denoise, and should thus receive more attention compared to non-critical tokens. 
We thus explicitly incorporate the timestep into the query mask prediction using AdaLN-single~\cite{Pixart-alpha}, enabling the network to better adapt to the denoising stage.
Additionally, given that query tokens are decoded sequentially across FFN blocks, we integrate the mask from the previous block into the query mask prediction of the current block. 
This helps maintain consistency in identifying critical tokens, facilitating more comprehensive denoising and improving the generation of complex details.
Specifically, as illustrated in~\autoref{fig:framework}, query mask prediction network (Q-MaskGen) takes the timestep $t$, cross-attention output $Z^{\text{fi}}_t$ and previous query mask $M^{i-1}_Q$ as the inputs, allowing prompt to better influence mask generation.
Q-MaskGen is composed of several MLP layers and can be expressed as follows:
\begin{equation}
    M^i_Q = \text{Q-MaskGen}(Z^{\text{fi}}_t, M^{i-1}_Q, t), i=1,...,N, M^0_q = I,
\end{equation}
where the $N$ denotes the number of FocusDiT blocks.

The query mask enables the model to suppress non-critical query tokens in the current timestep and block, ensuring that critical tokens make fuller use of the FFN to decode visual details.
Our default Q-Mask is a soft mask, and its hard version is only used as an optional thresholded form for inference acceleration.
Therefore, we apply it to the FFN output before the residual connection instead of inserting a hard mask between the two FFN linear layers.
This placement avoids disrupting the real-valued intermediate FFN representation while still selectively attenuating token-wise FFN updates.
In particular, as illustrated in~\autoref{fig:framework}, the input features $Z^{\text{fi}}_t$ are passed through the FFN to output updates, which are adaptively adjusted by the query mask $M_Q$ and added to the residual to produce the final result $Z^{\text{fo}}_t$. 
The calculation is as follows:
\begin{equation}
    \begin{aligned}
        Z^{\text{fo}}_t &= Z^{\text{fi}}_t + M_Q \cdot \text{FFN}(Z^{\text{fi}}_t) \\
                        &= Z^{\text{fi}}_t + M_Q \cdot [f(Z^{\text{fi}}_t\cdot {W_K} + b_K)\cdot W_{V} + b_V].
    \end{aligned}
    \label{eq:new_ffn}
\end{equation}
In fact, the query mask reweighs the influence of critical tokens on FFN vocabulary weight updates during backpropagation, preserving gradients from critical tokens to encourage the encoding of visual tokens corresponding to more complex visual details in the FFN vocabulary.
The backpropagation is conducted as follows:
\begin{equation}
    \begin{aligned}
    \frac{\partial \mathcal{L}_{DM}}{\partial W_K} &= M_Q \cdot \frac{\partial \mathcal{L}}{\partial Z^{\text{fo}}_t} \cdot \frac{\partial \text{FFN}(Z^{\text{fi}}_t)}{\partial W_K}, \\
    \frac{\partial \mathcal{L}_{DM}}{\partial W_V} &= M_Q \cdot \frac{\partial \mathcal{L}}{\partial Z^{\text{fo}}_t} \cdot \frac{\partial \text{FFN}(Z^{\text{fi}}_t)}{\partial W_V}.
    \end{aligned}
    \label{eq:ffn_grad}
\end{equation}



\subsection{Vocabulary Redistribution}
\label{sec:vr}
The FFN in DiT serves as vocabulary which provides query tokens with varying amounts of visual semantics embedded in visual tokens during generation. 
Therefore, designing a vocabulary size that aligns with token generation requirements can benefit the generation process. 
The utilization analysis in~\autoref{fig:insight} indicates that many vocabulary entries are redundant in some layers, but a manually designed layer-wise schedule can be brittle across model scales and training settings.
We therefore implement Vocabulary Redistribution (VR) with a local-expert and shared-expert FFN design.

For a standard FFN hidden size $C=4d$, where $d$ is the input feature dimension, we split the FFN vocabulary into a layer-specific local expert of size $C_1$ and a shared expert of size $C_2$, with $C_1+C_2=C$.
The local expert keeps layer-specific decoding capacity, while the shared expert stores reusable visual vocabulary shared across FocusDiT blocks.
For the $i$-th block, VR is computed as:
\begin{equation}
    \begin{aligned}
    \text{FFN}^{\text{VR}}_i(Z)
    &= f(Z W^{\text{loc}}_{K,i} + b^{\text{loc}}_{K,i}) W^{\text{loc}}_{V,i}
    + f(Z W^{\text{sha}}_{K} + b^{\text{sha}}_{K}) W^{\text{sha}}_{V}
    + b_{V,i}, \\
    W^{\text{loc}}_{K,i} &\in \mathbb{R}^{d \times C_1},\quad
    W^{\text{loc}}_{V,i} \in \mathbb{R}^{C_1 \times d}, \\
    W^{\text{sha}}_{K} &\in \mathbb{R}^{d \times C_2},\quad
    W^{\text{sha}}_{V} \in \mathbb{R}^{C_2 \times d}.
    \end{aligned}
\end{equation}
This design preserves the hidden size $C$ and thus the FFN computation of each block, but reduces parameters because the shared expert is reused instead of being independently instantiated in every block.
In our ablation, the shared expert size $C_2=3d$ gives the best tradeoff, as discussed in~\autoref{MaskGen}.



\section{Experiment}
\begin{table}[thpb]
    \centering
    \caption{Quantitative comparisons of text-to-image generation. Compared to PixArt-$\alpha$, our model uses less than 20\% of its computational resources, yet outperforms it in terms of FID and achieves comparable performance on CLIPScore.}
    \resizebox{\linewidth}{!}{
        \begin{tabular}{lccccc}
            \toprule
            Model               &   FID$\downarrow$        &   CLIPScore$\uparrow$ &  ImageReward$\uparrow$ & Training Time & Iterations \\
            \midrule
            Opensora-Plan        &    62.41              &  0.217   &   -1.85   &  --               & 346k \\
            PixArt-$\alpha$     &    30.50               &   0.315  &   0.74    &  753 A100 days    & --               \\
            SD3                 &   23.15                &  0.321   &   0.92    &   --              &  $\geq$ 500k           \\
            \midrule
            FocusDiT (Ours)     &     27.81              &   0.307  &   0.29    & \textbf{156 A100 days}     &  \textbf{72k}        \\
            \bottomrule
        \end{tabular}
    }
    \label{tab:my_label}
\end{table}

\begin{table*}[htpb]
    \centering
    \caption{Quantitative comparisons of text-to-image generation on GenEval benchmark.}
    \resizebox{0.98\linewidth}{!}{
        \begin{tabular}{lcccccccc}
         \toprule
        Models & Params. & \textbf{Overall $\uparrow$} & Single object & Two objects & Counting & Colors & Positions & Color Attribution \\
            \midrule
            minDALL-E~\cite{Minidalle3}           & 1.3B & \textbf{0.23} & 0.73 & 0.11 & 0.12 & 0.37 & 0.02 & 0.01  \\
            SD v1.5~\cite{SD}             & 0.9B & \textbf{0.43} & 0.97 & 0.38 & 0.35 & 0.76 & 0.04 & 0.06 \\
            PixArt-$\alpha$~\cite{Pixart-alpha}     & 0.6B & \textbf{0.48} & 0.98 & 0.50 & 0.44 & 0.80 & 0.08 & 0.07 \\
            SD v2.1~\cite{SD}             & 0.9B & \textbf{0.50} & 0.98 & 0.51 & 0.44 & 0.85 & 0.07 & 0.17 \\
            DALL-E 2~\cite{DellE2}            & 3.5B & \textbf{0.52} & 0.94 & 0.66 & 0.49 & 0.77 & 0.10 & 0.19 \\
            PixArt-$\Sigma$~\cite{Pixart-sigma}     & 0.6B & \textbf{0.52} & 0.98 & 0.59 & 0.50 & 0.80 & 0.10 & 0.15 \\
            SDXL~\cite{SDXL}                & 2.6B & \textbf{0.55} & 0.98 & 0.74 & 0.39 & 0.85 & 0.15 & 0.23 \\
            PlayGroundv2.5~\cite{PlayGroundv2.5}      & 2.6B & \textbf{0.56} & 0.98 & 0.77 & 0.52 & 0.84 & 0.11 & 0.17 \\
            SD3~\cite{SD3}                 & 2.1B & \textbf{0.70} & 0.99 & 0.88 & 0.60 & 0.85 & 0.30 & 0.59   \\
            \midrule
            Ours                & 1.0B & \textbf{0.57}  & 0.91    & 0.60 & 0.56 & 0.72 & 0.26 & 0.41  \\
            \bottomrule
        \end{tabular}
    }
    \label{tab:geneval-full}
\end{table*}



\subsection{Implementation Details}
We present the details of our  method, including datasets, model architecture, training and evaluation processes.

\noindent\textbf{Datasets.} 
Our model is trained on both real-world and synthesized text-image pairs. 
The real-world images are sourced from three datasets: Coyo-HD-11M~\cite{coyo}, LAION-Aesthetics-V1-120M~\cite{Laion} and LAION-Art-8M~\cite{Laion}, all chosen for their high aesthetic quality. 
The synthesized images come from Midjourney-Niji-1M~\cite{midjourney}, generated using Midjourney-V6.
Additionally, we expand the dataset with an internal collection of 500K text-image pairs.


\noindent\textbf{Training Details.}
We use Flan-T5-XXL~\cite{FlanT5XXL} for prompt encoding and VAE~\cite{Mochi} for image encoding, applying 8$\times$8 spatial downsampling while maintaining reconstruction quality.
The model is first trained from scratch on $256 \times 256$ resolution with 2048 batch size for diverse concepts learning, then on $512 \times 512$ with 1280 batch size for high-quality improvement. 
The training process takes around 60k steps, utilizing the AdamW~\cite{adamw} optimizer and a learning rate of $2 \times 10^{-4}$.

\noindent\textbf{Model Size and Inference Cost.}
The DiT baseline contains 1.078B parameters.
FocusDiT adds 0.199B parameters for MaskGen and reduces 0.254B parameters through the shared-FFN VR strategy, resulting in a final model size of 1.023B parameters.
For 20 denoising steps on an H800 GPU including VAE decoding, the baseline costs 5.9GB VRAM and 0.56s at 256px resolution, and 7.9GB VRAM and 0.72s at 512px resolution.
FocusDiT costs 5.8GB VRAM and 0.922s at 256px resolution, and 7.9GB VRAM and 0.92s at 512px resolution.

\noindent\textbf{Evaluation Metrics.} 
We assess the generation quality using four established metrics: Fréchet Inception Distance (FID)\cite{FID} measures the similarity between the feature distributions of generated and real images using a pre-trained Inception network, where a lower FID indicates a closer match to real images. CLIPScore\cite{CLIPScore} evaluates the semantic alignment between generated images and their corresponding prompts using the CLIP model~\cite{CLIP}, with higher scores indicating better alignment. GenEval\cite{GenEval} assesses the compositional quality of generated images, including object co-occurrence, spatial arrangement, object count, and color accuracy. ImageReward\cite{ImageReward} provides a perceptual evaluation of image quality based on human preference signals, capturing aspects such as aesthetics and coherence. 
In addition to these metrics, we introduce Structural Clarity and Textural Fidelity (SCTF) to assess the perceptual quality of generated images. SCTF is measured using Vision-Language Models (VLMs) such as Qwen2.5-VL\cite{qwen2.5} and InternVL-2.5\cite{internvl2.5}, which assign a score from 0 to 10 based on the clarity of structural elements and the fidelity of textures, ensuring the absence of artifacts or hallucinations. We use SCTF as a complementary perceptual metric rather than a replacement for standard distributional, alignment, and human-preference evaluations. Specifically, for each model and configuration, we compute the \textbf{SCTF score} as the \textbf{mean score} across all generated images:  

\begin{equation}
SCTF = \frac{1}{N} \sum_{i=1}^{N} s_i
\end{equation}
where \( N \) is the total number of images, and \( s_i \) is the SCTF score assigned to the \( i \)-th image by the VLM.  
To quantify the variability of SCTF scores, we report the \textbf{standard error of the mean (SEM)}, which is computed as:  
\begin{equation}
SEM = \frac{\sigma}{\sqrt{N}}
\end{equation}
where \( \sigma \) is the standard deviation of the SCTF scores across all images. The SEM provides an estimate of the uncertainty in the mean SCTF score, allowing for more reliable comparisons between different model configurations. In all experiments, we generated a total of 5,120 images for evaluation, i.e., $N=5120$.
SCTF provides a fine-grained evaluation of image quality beyond traditional statistical measures.


\subsection{Qualitative Analysis}
In comparison to other DiT-based models such as PixArt-$\alpha$, SD3, and OpenSoraPlan~\cite{opensoraplan}, our proposed FocusDiT demonstrates the superiority in various aspects of image generation as shown in ~\autoref{fig:quali}. 
FocusDiT excels in generating finer details, especially in character features like hair textures, which are more intricate and realistic compared to the other models. 
It also produces more structurally sound and coherent architectural designs, maintaining better proportions and plausibility in generated buildings. 
Additionally, when tasked with generating non-realistic or imaginative scenes, such as "a kitten dressed in judo uniform," FocusDiT delivers results with superior structural refinement and more nuanced details, effectively capturing the complexity of abstract concepts. 
These results underline the strengths of FocusDiT in both realistic and creative image generation, offering enhanced fidelity in character details and structural accuracy in complex environments.

\noindent\textbf{Query Masks Visualization.}
As shown in ~\autoref{fig:mq_vis}, we visualize the Query Masks predicted by our Q-MaskGen across each block through 30 denoising timesteps. 
From the perspective in the FocusDiT blocks, the query masks are predicted from the first block to the last sequentially, where we observe that both shallow and deep blocks tend to mask a significant portion of the queries representing high-frequency details such as car edges. 
Specifically, in the shallowest and deepest blocks, the query mask effectively masks out almost all query tokens during the FFN update, indicating that decoding these tokens does not rely on these FFNs.
Based on this observation, we are able to accelerate the inference process by skipping layers with query masks close to zero, as discussed in \autoref{sec:acc}.
From the denoising timestep perspective, query tokens gradually transition from Gaussian noise to a structured image, and the contours of the foreground and background become increasingly refined as denoising progresses. 
Notably, we also observe that the selected critical token does not continuously focus on the generation of fine details in the subject at different timesteps. 
Timesteps corresponding to high noise level are more focused on generating the background details, while timesteps with lower noise shift their attention toward the foreground and finer structures, as evidenced in ~\autoref{fig:mq_vis}.

\begin{figure}[htpb]
    \centering
    \includegraphics[width=\linewidth]{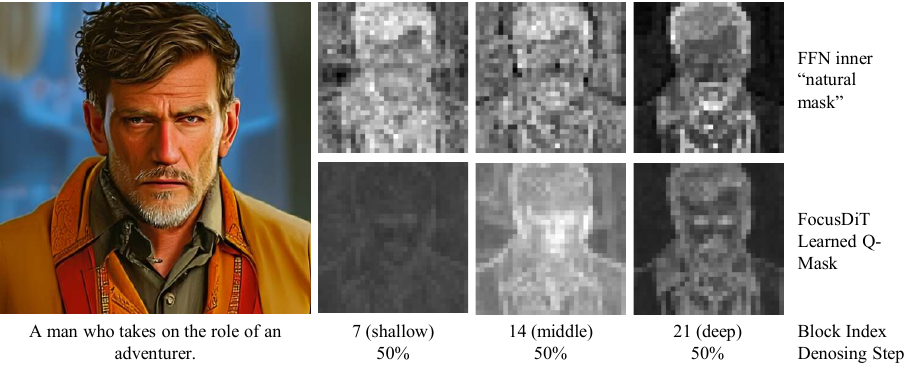}
    \caption{Comparison between the implicit FFN natural mask and our learned Q-Mask.}
    \label{fig:natural-mask}
\end{figure}
\begin{figure*}[htpb]
    \centering
    \includegraphics[width=0.8\linewidth]{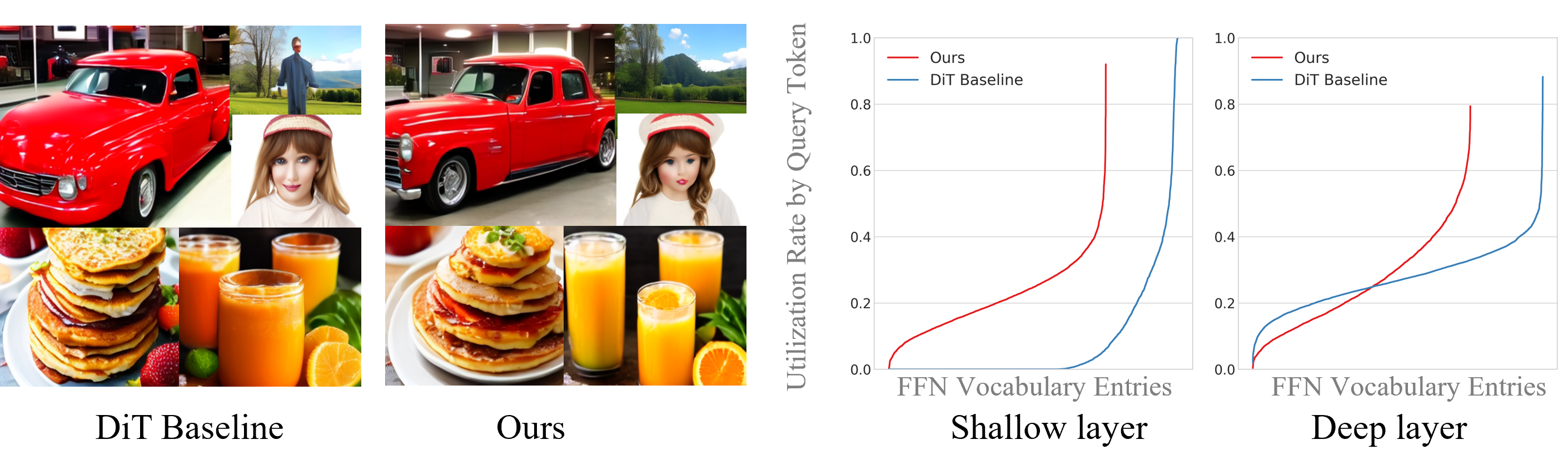}
    \caption{
    Comparison of vocabulary utilization between the DiT baseline and FocusDiT across various prompts.
    }
    \label{fig:fail}
\end{figure*}
\noindent\textbf{Natural Mask and Learned Q-Mask.}
The activation pattern inside a standard FFN can be viewed as an implicit ``natural mask'', because only positively activated vocabulary entries contribute to each query token after the FFN activation.
As shown in~\autoref{fig:natural-mask}, this natural mask contains weak semantic clustering but is often noisy and spatially chaotic.
In contrast, the Q-Mask learned by MaskGen is smoother and more semantically coherent.
In shallow and deep blocks, it highlights edge, structure, and texture regions; in middle blocks, it more clearly separates facial, clothing, and background regions.
This comparison indicates that FocusDiT does not merely rely on the incidental FFN activation pattern, but explicitly learns a more interpretable focus mechanism for token decoding.

\noindent\textbf{Vocabulary Utilization.}
As shown in \autoref{fig:fail}, we visualize and compare the utilization rates of query tokens at different layers between FocusDiT and DiT. 
In shallow blocks, FocusDiT demonstrates a significantly higher utilization rate compared to DiT, which can be attributed to our approach of redistributing vocabulary capacity across the layers. 
For the deep layers, FocusDiT demonstrates a relatively higher utilization rate.
This redistribution allows FocusDiT to make more efficient use of the entire vocabulary.

\subsection{Quantitative Evaluation}
\label{quant}
From the quantitative analysis in \autoref{tab:my_label}, it is observed that although the foundational architecture of FocusDiT is quite similar to that of OpenSoraPlan and PixArt-alpha, our model significantly outperforms OpenSoraPlan and achieves a better FID than PixArt-alpha. Notably, this achievement comes despite PixArt-alpha being trained on a large amount of internal data, whereas FocusDiT primarily relies on open-source datasets. 
However, the limited scale and diversity of available open-source data have prevented our model from surpassing SD3. We believe that with access to larger and more diverse datasets and additional training time, FocusDiT’s performance could be further enhanced to close the remaining gap with SD3 in future work.
The detailed GenEval comparison in~\autoref{tab:geneval-full} further shows that FocusDiT reaches an overall score of 0.57, outperforming several public text-to-image models while using a relatively moderate model size and training cost.

\begin{figure}[htpb]
    \centering
    \includegraphics[width=\linewidth]{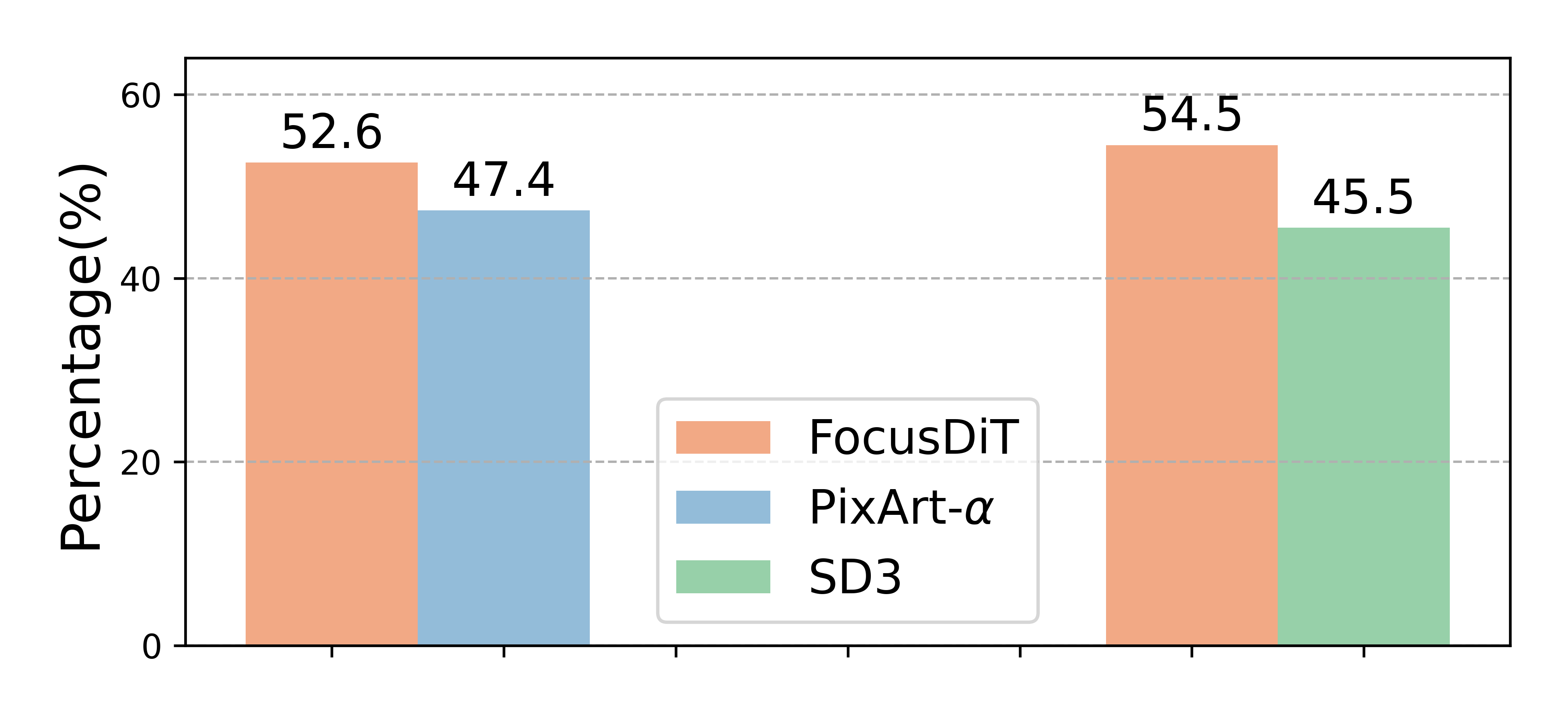}
    \caption{User Study. Our FocusDiT outperforms both PixArt-$\alpha$ and SD3 in human preference evaluation.}
    \label{fig:user}
\end{figure}

\noindent\textbf{User Study.} We conduct a user study for evaluating fine-grained structure in image generation, which compares the performance of our image generation method with two competing models: PixArt-$\alpha$~\cite{Pixart-alpha} and SD3~\cite{SD3}.
After 660 pairwise user comparisons, the results presented in~\autoref{fig:user} clearly demonstrate that our proposed method outperforms both PixArt-$\alpha$ and SD3 in human preference quality. 
Specifically, compared to PixArt-$\alpha$, our method achieves a 52.6\% win rate, and when compared to SD3, our method surpasses it with a 54.5\% win rate. 
This suggests that our image generation method not only matches but also outperforms the other models for less artifacts and better fine-grained details with the given prompts.

\subsection{Ablation Study}
\begin{table*}[t]
    \centering
    \caption{Ablation studies of FocusDiT at 30k training steps. The upper table evaluates MaskGen design choices with full metrics; the lower table reports the focused shared-expert VR sweep using FID as the selection criterion.}
    \resizebox{0.98\linewidth}{!}{
        \begin{tabular}{@{}c l c c c | c c c c@{}}
            \toprule
            \multicolumn{2}{c}{Models} & \multicolumn{3}{c|}{Image Quality} & \multicolumn{4}{c}{Structural Clarity and Textural Fidelity (SCTF)$\uparrow$} \\
            \cmidrule(r){3-5} \cmidrule(l){6-9}
            & & FID$\downarrow$ & CLIPScore$\uparrow$ & ImageReward$\uparrow$ & InternVL2.5-4B & InternVL2.5-8B & Qwen2.5-VL-3B & Qwen2.5-VL-7B \\
            \midrule
            A & DiT Baseline & 31.29  & 0.310 &  -0.71 & 7.5959{\scriptsize ±0.0208} & 8.1184{\scriptsize ±0.0119} & 6.4705{\scriptsize ±0.0193} & 8.1785{\scriptsize ±0.0126} \\
            \midrule
            $\text{B}_1$ & A + 2layers & 33.59 & 0.306 & -0.87 & 7.4889{\scriptsize ±0.0224} & 8.0793{\scriptsize ±0.0129} & 6.5273{\scriptsize ±0.0198} & 8.1488{\scriptsize ±0.0134} \\
            $\text{B}_2$ & A + 5layers & 31.42 & 0.314 & -0.66 & 7.7328{\scriptsize ±0.0195} & 8.2113{\scriptsize ±0.0106} & 6.5854{\scriptsize ±0.0201} & 8.2137{\scriptsize ±0.0127} \\
            $\text{B}_3$ & A + 10layers & 31.58 & 0.312 & -0.65 & 7.7705{\scriptsize ±0.0172} & 8.2203{\scriptsize ±0.0099} & 6.5744{\scriptsize ±0.0187} & 8.2480{\scriptsize ±0.0103} \\
            \midrule
            C & $\text{B}_2$ + Dropout & 31.00 & 0.311 & -0.70 & 7.6693{\scriptsize ±0.0195} & 8.1773{\scriptsize ±0.0111} & 6.5008{\scriptsize ±0.0197} & 8.1625{\scriptsize ±0.0124} \\
            D & $\text{B}_2$ + Premask & 32.19 & 0.315 & -0.57 & 7.7668{\scriptsize ±0.0175} & 8.2256{\scriptsize ±0.0098} & 6.6541{\scriptsize ±0.0181} & 8.2824{\scriptsize ±0.0110} \\  
            \bottomrule
        \end{tabular}
    }
    \vspace{0.6em}

    \resizebox{0.46\linewidth}{!}{
        \begin{tabular}{@{}l c c@{}}
            \toprule
            VR variant & Shared expert size & FID$\downarrow$ \\
            \midrule
            C + D + VR-shared-$d$  & $d$  & 32.33 \\
            C + D + VR-shared-$2d$ & $2d$ & 30.59 \\
            C + D + VR-shared-$3d$ & $3d$ & \textbf{30.58} \\
            \bottomrule
        \end{tabular}
    }
    \label{MaskGen}
\end{table*}

\begin{figure}[htpb]
    \centering

    \includegraphics[width=\linewidth]{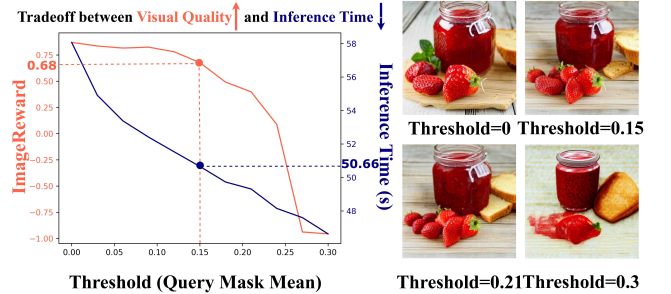}
    \caption{Inference time reduction by skipping FFN calculation guided by the mean of query masks.}
    \label{fig:acce}
\end{figure}
\section{Inference Acceleration}
\label{sec:acc}

To prevent unnecessary overhead due to the increase in model parameters, we adopted a simple Multi-Layer Perceptron (MLP) structure for MaskGen. We conducted carefully designed ablation experiments to demonstrate the effectiveness of our architecture.
First, we assessed the impact of the number of MLP layers by testing 2-layer, 5-layer, and 10-layer configurations, with the 2-layer and 10-layer networks representing shallower and deeper designs, respectively. The 5-layer MLP achieved the best balance between model complexity and performance, as shown in~\autoref{MaskGen}.
Next, adding a Dropout layer encouraged the model to mask out more queries during training, enhancing focus on critical queries and refining the mask generation process.
Additionally, we also introduced a prior mask from the previous layer into MaskGen. This enhancement allowed the model to better concentrate on previously masked regions, resulting in higher-quality mask generation and improved final image generation.
For VR, we ablated the shared-expert size under the same 30k training steps and batch size of 1024.
The shared-$d$, shared-$2d$, and shared-$3d$ variants achieve FIDs of 32.33, 30.59, and 30.58, respectively, compared with 31.29 for the DiT baseline.
Since this sweep is used only to select the shared-expert size, we report FID rather than repeating the full metric suite used for MaskGen ablations.
We therefore adopt shared-$3d$ as the default VR design.
After extending the shared-$3d$ model to 50k training steps, FocusDiT obtains 0.57 on GenEval, as reported in~\autoref{tab:geneval-full}.
Through these ablation experiments, we demonstrate that each design choice in our MaskGen module contributes significantly to the overall performance of FocusDiT, ensuring both efficient and high-quality image generation.

The query mask reflects how much a token relies on the FFN's vocabulary. 
We believe that the lower the mean of the query mask, the smaller the contribution of the current FFN to token denoising, therefore the model can skip these FFNs to achieve accelerated inference while preserving its overall performance. 
We analyze the tradeoff between the model's overall performance and inference efficiency by setting a threshold. 
As shown in ~\autoref{fig:acce}, when the threshold is low, fewer FFNs are skipped, and the model's overall performance is preserved as much as possible; when the threshold is high, more FFNs are skipped, which significantly improves the model's inference efficiency, but severely impacts its overall performance.
Besides, we observe that lower thresholds have little impact on network performance, but once the threshold exceeds a certain range, the model's performance drops significantly. 
With a threshold of 0.15, the time to generate 16 samples decreases from 58.69s to 50.66s without significant performance drop.

\begin{figure}[htpb]
    \centering

    \includegraphics[width=\linewidth]{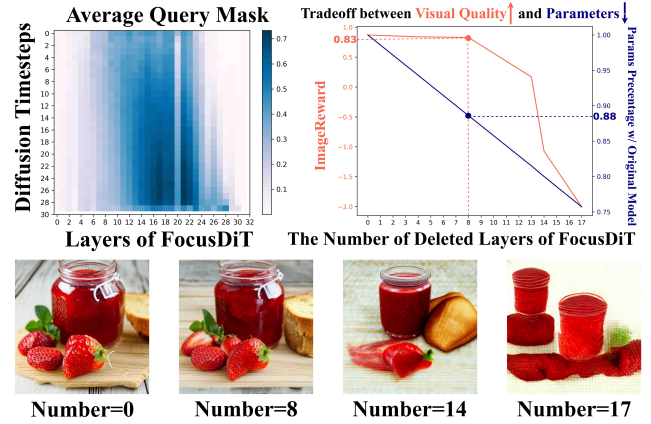}
    \caption{Parameters reduction by removing FFN guided by the mean of query mask.}
    \label{fig:acce2}
\end{figure}


\noindent\textbf{Parameter Reduction.}
To further explore the impact of the less important FFN on token denoising, as shown in ~\autoref{fig:acce2}, we calculate the average of the mean values of query masks across 200 different prompts, at various timesteps and different layers of FocusDiT.
We find that the mean value of the query mask remains relatively consistent across different timesteps. However, there is a notable variation in this value across different layers of FocusDiT.
Based on this observation, we instruct the model to skip specific FFNs across all timesteps, in other words, we prune the model's parameter by removing some of the FFN parameters in certain layers.
As shown in ~\autoref{fig:acce2}, as the number of deleted layers of FocusDiT increases, the overall performance of the model initially shows no significant change, then suddenly decreases substantially.
After removing 8 relatively unimportant FFNs, the model's parameter count is reduced to 88\% of the original, while the overall performance of the model shows no significant degradation.

\section{Conclusion}
\label{sec:conclusion}
In this paper, we introduce a Query Token Masking strategy aimed at enhancing the utilization of critical query tokens with complex structures, allowing the model to capture finer details in visual generation tasks.
Additionally, we propose a Vocabulary Redistribution (VR) scheme that reallocates vocabulary capacity across network layers, ensuring better alignment with the specific needs of these query tokens. 
This approach not only improves vocabulary efficiency but also reduces inference costs, ultimately leading to more accurate and computationally efficient models for image generation, particularly in fine-grained details.
Furthermore, we also introduce a threshold-based method that skips certain Feed-Forward Networks based on the query mask, balancing inference speed and performance.

\newpage
{
    \small
    \bibliographystyle{ACM-Reference-Format}
    \bibliography{main}
}


\end{document}